\documentclass[conference,letter]{IEEEtran}  
\IEEEoverridecommandlockouts
\usepackage{amsmath,amsfonts}
\usepackage{algorithmic}
\usepackage{algorithm}
\usepackage{array}
\usepackage[caption=false,font=normalsize,labelfont=sf,textfont=sf]{subfig}
\usepackage{textcomp}
\usepackage{stfloats} 
\usepackage{url}
\usepackage{verbatim}
\usepackage{graphicx}
\usepackage{cite}
\usepackage{subcaption}
\DeclareMathOperator*{\argmax}{argmax}
\usepackage{amssymb}
\usepackage{orcidlink}
\usepackage{overpic}
\usepackage{float}
\usepackage{graphicx} 
\usepackage{tabularx}
\usepackage{longtable}
\usepackage{booktabs}
\usepackage{lscape}
\usepackage{multirow}
\usepackage{multicol}

\usepackage{tikz}

\usepackage[dvipsnames,svgnames,table, xcdraw]{xcolor}

\usetikzlibrary{arrows.meta, fit, positioning, backgrounds, calc}
\usepackage{pgfplots}
\pgfplotsset{compat=1.18}

\definecolor{Cactor}{RGB}{30, 100, 185}
\definecolor{Ccritic}{RGB}{195, 95, 15}
\definecolor{Cenv}{RGB}{28, 135, 65}
\definecolor{Cbuf}{RGB}{105, 55, 175}
\definecolor{Calgo}{RGB}{55, 55, 65}
\definecolor{Cpd}{RGB}{175, 40, 85}
\definecolor{Cpanel}{RGB}{228, 238, 252}
\definecolor{Cpborder}{RGB}{155, 185, 225}
\definecolor{buildingColor}{rgb}{0.5020, 0, 0}
\definecolor{vegetationColor}{rgb}{0.5020, 0.5020, 0}
\definecolor{carColor}{rgb}{0.2510, 0, 0.5020}
\definecolor{pedestrainColor}{rgb}{0.2510, 0.2510, 0}
\definecolor{roadColor}{rgb}{0.5020, 0.2510, 0.5020}
\definecolor{fenceColor}{rgb}{0.2510, 0.2510, 0.5020}
\definecolor{signateColor}{rgb}{0.7529, 0.5020, 0.5020}
\definecolor{sidewalkColor}{rgb}{0, 0, 0.7529}
\definecolor{poleColor}{rgb}{0.7529, 0.7529, 0.5020}

\definecolor{nwaterColor}{rgb}{0.1176, 0.5647, 0.9804}
\definecolor{nroadColor}{rgb}{0.9804, 0.9804, 0.9804}
\definecolor{nsidewalkColor}{rgb}{0.5020, 0.2510, 0.5020}
\definecolor{nterrainColor}{rgb}{0.5020, 0.5020, 0}
\definecolor{nbuildingColor}{rgb}{0.9804, 0.5020, 0}
\definecolor{nvegetationColor}{rgb}{0.4196, 0.5569, 0.1373}
\definecolor{ncarColor}{rgb}{0, 0, 0.5569}
\definecolor{npersonColor}{rgb}{0.8627, 0.0784, 0.2353}
\definecolor{nbikeColor}{rgb}{0.4667, 0.0431, 0.1255}
\definecolor{npoleColor}{rgb}{0.7529, 0.7529, 0.7529}
\definecolor{nstairColor}{rgb}{0.4824, 0.4078, 0.9333}
\definecolor{ntrafficsignColor}{rgb}{0.9804, 0.9804, 0}

\definecolor{scarColor}{rgb}{0.9608, 0.5882, 0.3922}
\definecolor{sbicycleColor}{rgb}{0.9608, 0.9020, 0.3922}
\definecolor{smotorcycleColor}{rgb}{0.5882, 0.2353, 0.1176}
\definecolor{struckColor}{rgb}{0.7059, 0.1176, 0.3137}
\definecolor{sothervehicleColor}{rgb}{1, 0.3137, 0.3922}
\definecolor{spersonColor}{rgb}{0.1176, 0.1176, 1}
\definecolor{sbicyclistColor}{rgb}{0.7843, 0.1569, 1}
\definecolor{smotorcyclistColor}{rgb}{0.3529, 0.1176, 0.5882}
\definecolor{sroadColor}{rgb}{1, 0, 1}
\definecolor{sparkingColor}{rgb}{1, 0.5882, 1}
\definecolor{ssidewalkColor}{rgb}{0.2941, 0, 0.2941}
\definecolor{sothergroundColor}{rgb}{0.2941, 0, 0.6863}
\definecolor{sbuildingColor}{rgb}{0, 0.7843, 1}
\definecolor{sfenceColor}{rgb}{0.1961, 0.4706, 1}
\definecolor{svegetationColor}{rgb}{0, 0.6863 ,0}
\definecolor{strunkColor}{rgb}{0, 0.2353, 0.5294}
\definecolor{sterrainColor}{rgb}{0.3137, 0.9412, 0.5882}
\definecolor{spoleColor}{rgb}{0.5882, 0.9412, 1}
\definecolor{strafficsignColor}{rgb}{0, 0, 1}

\definecolor{beamColor}{rgb}{0.9608, 0.5882, 0.3922}
\definecolor{boardColor}{rgb}{0.9608, 0.9020, 0.3922}
\definecolor{bookcaseColor}{rgb}{0.5882, 0.2353, 0.1176}
\definecolor{ceilingColor}{rgb}{0.7059, 0.1176, 0.3137}
\definecolor{chairColor}{rgb}{1, 0.3137, 0.3922}
\definecolor{clutterColor}{rgb}{0.1176, 0.1176, 1}
\definecolor{doorColor}{rgb}{0.7843, 0.1569, 1}
\definecolor{floorColor}{rgb}{0.3529, 0.1176, 0.5882}
\definecolor{tableColor}{rgb}{1, 0, 1}
\definecolor{wallColor}{rgb}{1, 0.5882, 1}

\newcommand{\best}[1]{\textbf{\textcolor[HTML]{B0006C}{#1}}}    
\newcommand{\second}[1]{\underline{#1}}                            

\begin{document}

\title{Efficient Continuous Semantic Mapping based on Spatio-Temporal Awareness}

\author{
\IEEEauthorblockN{My Le Pham\textsuperscript{1}, Dinh Trieu Duong\textsuperscript{1}, $^*$Xiem HoangVan\textsuperscript{1}\orcidlink{0000-0002-7524-6529}, and Thanh Nguyen Canh\textsuperscript{1,2}~\orcidlink{0000-0001-6332-1002}}
\IEEEauthorblockA{\textsuperscript{1}University of Engineering and Technology, Vietnam National University, 10000, Hanoi, Vietnam. 
}
\IEEEauthorblockA{\textsuperscript{2}School of Information Science, Japan Advanced Institute of Science and Technology, Nomi, 923-1211, Ishikawa, Japan. 
}
\IEEEauthorblockA{*Corresponding author: \tt xiemhoang@vnu.edu.vn}
}


\maketitle

\makeatletter
\setlength{\@fptop}{0pt}
\makeatother

\begin{abstract}
Continuous semantic mapping allows autonomous robots to understand both the spatial structure and the semantic content of complex environments. However, most existing methods process the entire space, treat voxels as independent units, and do not keep the semantic labels consistent over time. This leads to high computational cost and reduced robustness in dynamic scenes. This paper proposes a semantic mapping method that brings spatial and temporal relationships into the semantic inference process. The method adjusts the inference range according to the local semantic uncertainty and fuses labels over time to improve map stability and computational efficiency. Experiments on the SemanticKITTI dataset show that the proposed method improves mapping accuracy by about 12\% and reaches an mIoU of 54.92\%, which is 13.18 percentage points higher than spatial-only mapping. These results show that spatio-temporal reasoning is effective for continuous semantic mapping in autonomous robotic systems.
\end{abstract}

\begin{IEEEkeywords}
Continuous semantic mapping, spatio-temporal inference, semantic uncertainty, Bayesian kernel inference.
\end{IEEEkeywords}

\section{Introduction}
\label{sec:introduction}

Autonomous robots are now used in self-driving vehicles, service robots, rescue robots, and other intelligent systems~\cite{chen2025survey}. To act safely, a robot must localize itself and build a map of its surroundings at the same time, which is the role of simultaneous localization and mapping (SLAM)~\cite{cadena2016past}. Traditional SLAM builds geometric maps, such as occupancy grids, point clouds, or 3D voxel maps~\cite{hornung2013octomap, campos2021orb}. These maps show where obstacles are, but not what they are: a geometric map cannot tell whether an obstacle is a vehicle, a pedestrian, a tree, or a building. Semantic mapping removes this limitation by adding semantic labels to the geometric map~\cite{canh2026semantic, canh2023object}, so the robot can understand the scene in a way that is closer to human perception and can support context-aware navigation, obstacle avoidance, and planning.

Building accurate and stable semantic maps is still difficult. Semantic labels are usually produced by deep-learning segmentation models, which are affected by sensor noise, occlusion, sparse points, and similar-looking classes~\cite{deng2024seecsom, canh2024s3m}. When wrong labels are projected into the map, they add local noise and reduce its consistency. In addition, the same region is observed many times as the robot moves, and changes in viewpoint and sensing can assign the same voxel different labels at different times. Many existing methods treat voxels as independent units~\cite{doherty2017bayesian, zhi2019continuous, gan2020bayesian} or model only the spatial relationship between neighbors~\cite{wilson2024convbki}, so this temporal information is not used, and the map suffers from semantic drift.

To address these problems, we propose a continuous semantic mapping method based on spatio-temporal inference. The method measures the semantic uncertainty of each voxel and uses it to adapt the kernel range during spatial inference: uncertain regions use a larger range to gather more context, while confident regions use a smaller range to save computation. A temporal decay mechanism then fuses past and current observations and gives more weight to recent ones, which keeps the map consistent over time. On the SemanticKITTI dataset~\cite{behley2019semantickitti}, the method reaches an mIoU of 54.92\% and improves mapping accuracy by about 12\% over spatial-only mapping.
The main contributions of this paper are summarized as follows:
\begin{itemize}
\item We propose a continuous semantic mapping framework that jointly exploits spatial and temporal information for outdoor LiDAR scenes.
\item We use semantic uncertainty to adapt the kernel length, which balances mapping accuracy and computational cost.
\item We introduce a time-aware temporal fusion mechanism that reduces semantic drift and improves long-term map stability.
\end{itemize}

The remainder of this paper is organized as follows. Section~\ref{sec:related} reviews related work. Section~\ref{sec:method} describes the proposed system for efficient continuous semantic mapping with Bayesian kernel inference. Section~\ref{sec:exp} presents the experiments and analyzes the results. Section~\ref{sec:conclusion} concludes the paper and discusses limitations and future work.
\section{Related Work}
\label{sec:related}


\subsection{Continuous Semantic Mapping}
Continuous mapping methods relax the assumption that map cells are independent. Hilbert maps~\cite{zhi2019continuous} represent occupancy as a continuous function in a high-dimensional feature space, which lowers the memory cost in large environments. Bayesian kernel inference~\cite{doherty2017bayesian} models the spatial correlation between points with kernel functions and was later extended to dense semantic mapping~\cite{gan2020bayesian}. More recently, SEE-CSOM~\cite{deng2024seecsom} uses a kernel length that adapts to semantic uncertainty in order to control how far information spreads in space, and ConvBKI~\cite{wilson2024convbki} performs Bayesian kernel inference inside a depthwise convolution layer, which keeps the probabilistic and interpretable nature of kernel methods while running in real time. These methods produce smooth maps with sharp boundaries, but they mainly model spatial relationships and do not fully use the temporal correlation between consecutive scans.

\subsection{Spatio-Temporal Semantic Mapping}
A few methods add a temporal dimension. CT-Map~\cite{zeng2018semantic} treats object classes as hidden states in a dynamic probabilistic model and updates both pose and label over time, but it works at the object level in a discrete space. STAMICS~\cite{yi2026stamics} adds temporally consistent semantics to dense RGB-D reconstruction, but its main goal is camera tracking and indoor scene reconstruction rather than continuous outdoor mapping. Khronos~\cite{schmid2024khronos} builds a dense spatio-temporal metric-semantic map that reasons about both short-term dynamics and long-term changes, but it focuses on object-level change detection rather than continuous voxel-level semantic inference. A parallel line of work represents the scene with neural implicit fields or 3D Gaussians, such as SNI-SLAM~\cite{zhu2024sni} and SGS-SLAM~\cite{li2024sgs}, which give dense and photo-realistic semantic reconstructions but mainly target indoor RGB-D scenes and need heavy optimization. 

In general, existing methods model either spatial correlation or temporal correlation, but they rarely propagate and accumulate semantic uncertainty in both space and time within a single continuous framework. Our method fills this gap by combining adaptive spatial kernel inference with a time-aware fusion mechanism, which keeps the semantic map both locally smooth and temporally consistent.

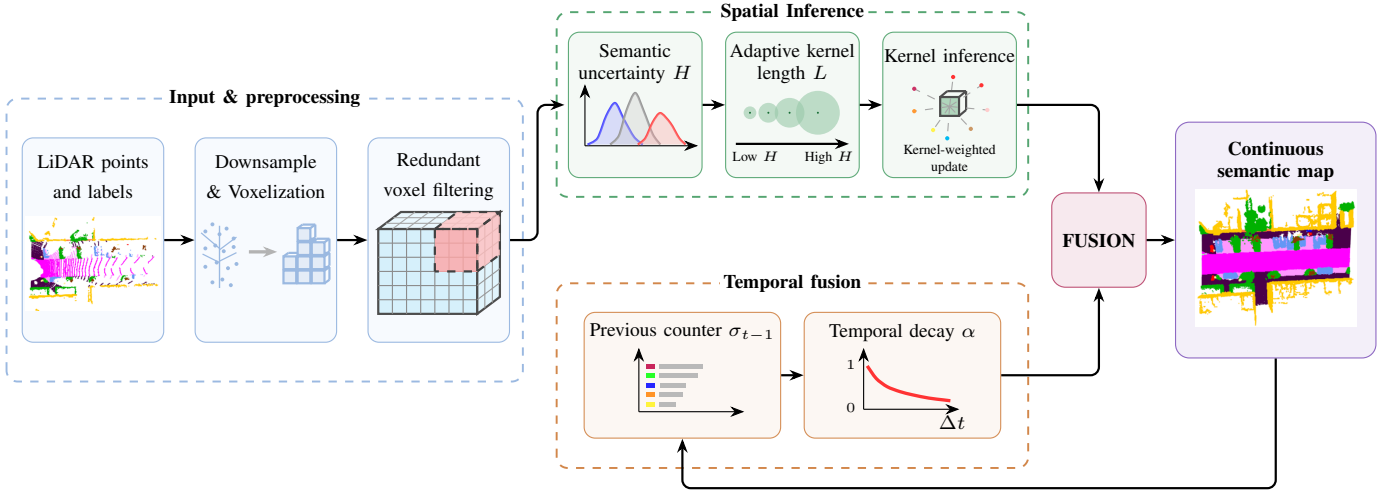
\begin{figure*}[!ht]
\centering
\resizebox{\textwidth}{!}{%
\begin{tikzpicture}[
    font=\scriptsize,
    >={Stealth[length=1.8mm]},
    arr/.style={->, thick},
    paneltitle/.style={font=\bfseries\scriptsize, fill=white, inner sep=1pt},
    boxtext/.style={font=\scriptsize, align=center},
    smalltext/.style={font=\tiny, align=center}
]

\draw[dashed, rounded corners=4pt, draw=Cpborder, line width=0.7pt]
    (-7.4,-1.8) rectangle (-0.83,1.8);
\node[paneltitle] at (-4.115,1.8) {Input \& preprocessing};

\draw[dashed, rounded corners=4pt, draw=Cenv!75, line width=0.7pt]
    (-0.4,0.55) rectangle (5.6,2.9);
\node[paneltitle] at (2.6,2.9) {Spatial Inference};

\draw[dashed, rounded corners=4pt, draw=Ccritic!75, line width=0.7pt]
    (-0.4,-2.9) rectangle (5.6,-0.55);
\node[paneltitle] at (2.6,-0.55) {Temporal fusion};


\node[
    draw=Cpborder,
    rounded corners=4pt,
    fill=Cpanel!30,
    minimum width=1.8cm,
    minimum height=2.6cm,
    inner sep=1mm
] (lidar) at (-6.3,0) {};

\node[boxtext] at ($(lidar.north)+(0,-0.5)$)
{LiDAR points\\[1.mm] and labels};

\node at ($(lidar.center)+(0,-0.3)$)
{\includegraphics[width=1.72cm]{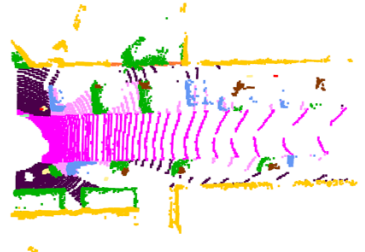}};

\node[
    draw=Cpborder,
    rounded corners=4pt,
    fill=Cpanel!30,
    minimum width=1.8cm,
    minimum height=2.6cm,
    inner sep=1mm
] (down) at (-4.1,0) {};

\node[boxtext] at ($(down.north)+(0,-0.5)$){Downsample\\[1mm] \& Voxelization};

\begin{scope}[shift={($(down.center)+(0.13,-0.25)$)}, scale=0.18]
    \foreach \x/\y in {
        -5/2,-4.5/2.5,-4/1.8,-3.5/2.2,-3/1.5,
        -5/0.8,-4.3/1.0,-3.7/0.6,-3/0.9,
        -5/-0.5,-4.2/-0.2,-3.4/-0.6,-2.8/-0.3,
        -4.5/-1.6,-3.8/-1.4,-3.2/-1.8}
        \fill[Cpborder] (\x,\y) circle (0.14);

    \draw[Cpborder, line width=0.5pt] (-4,-2.0) -- (-4,2.4);
    \draw[Cpborder, line width=0.45pt] (-4,0.8) -- (-5.3,1.9);
    \draw[Cpborder, line width=0.45pt] (-4,0.7) -- (-2.8,1.6);
    \draw[Cpborder, line width=0.45pt] (-4,-0.2) -- (-5.2,0.4);
    \draw[Cpborder, line width=0.45pt] (-4,-0.1) -- (-2.8,0.5);
\end{scope}

\begin{scope}[shift={($(down.center)+(0,-0.15)$)}, scale=0.18]
    \draw[->, very thick, gray!60] (-1.2,0) -- (0.8,0);
\end{scope}

\begin{scope}[shift={($(down.center)+(-0.1,-0.55)$)}, scale=0.18]
    \foreach \x/\y in {
        1.8/0,2.8/0,3.8/0,
        1.8/1,2.8/1,3.8/1,
        2.8/2,3.8/2,
        2.8/3}
    {
        \draw[Cpborder, thick, fill=Cactor!12] (\x,\y) rectangle ++(0.9,0.9);
        \draw[Cpborder, thick] (\x+0.9,\y) -- ++(0.35,0.25);
        \draw[Cpborder, thick] (\x+0.9,\y+0.9) -- ++(0.35,0.25);
        \draw[Cpborder, thick] (\x,\y+0.9) -- ++(0.35,0.25);
        \draw[Cpborder, thick] (\x+0.35,\y+1.15) -- ++(0.9,0);
        \draw[Cpborder, thick] (\x+1.25,\y+0.25) -- ++(0,0.9);
    }
\end{scope}

\node[
    draw=Cpborder,
    rounded corners=4pt,
    fill=Cpanel!30,
    minimum width=1.8cm,
    minimum height=2.6cm,
    inner sep=1mm
] (rvfm) at (-1.9,0) {};

\node[boxtext] at ($(rvfm.north)+(0,-0.5)$)
{Redundant\\[1mm] voxel filtering};
\begin{scope}[shift={($(rvfm.center)+(-0.05,-0.399)$)}, scale=0.18]

    \def\xmin{-4}
    \def\xmax{3}
    \def\ymin{-3}
    \def\ymax{3}
    \def\dx{1.6}
    \def\dy{1.2}


    \path[fill=cyan!14] (\xmin,\ymin) rectangle (\xmax,\ymax);
    \draw[black!70, thick] (\xmin,\ymin) rectangle (\xmax,\ymax);

    \foreach \x in {-3,-2,-1,0,1,2}
        \draw[black!45] (\x,\ymin) -- (\x,\ymax);
    \foreach \y in {-2,-1,0,1,2}
        \draw[black!45] (\xmin,\y) -- (\xmax,\y);

    \path[fill=cyan!10]
        (\xmax,\ymin) -- (\xmax+\dx,\ymin+\dy) --
        (\xmax+\dx,\ymax+\dy) -- (\xmax,\ymax) -- cycle;
    \draw[black!70, thick]
        (\xmax,\ymin) -- (\xmax+\dx,\ymin+\dy) --
        (\xmax+\dx,\ymax+\dy) -- (\xmax,\ymax) -- cycle;

    \foreach \y in {-2,-1,0,1,2}
        \draw[black!40] (\xmax,\y) -- (\xmax+\dx,\y+\dy);

    \foreach \t in {1,2}
        \draw[black!40]
        (\xmax+\t*\dx/3,\ymin+\t*\dy/3) --
        (\xmax+\t*\dx/3,\ymax+\t*\dy/3);

    \path[fill=cyan!7]
        (\xmin,\ymax) -- (\xmin+\dx,\ymax+\dy) --
        (\xmax+\dx,\ymax+\dy) -- (\xmax,\ymax) -- cycle;
    \draw[black!70, thick]
        (\xmin,\ymax) -- (\xmin+\dx,\ymax+\dy) --
        (\xmax+\dx,\ymax+\dy) -- (\xmax,\ymax) -- cycle;

    \foreach \x in {-3,-2,-1,0,1,2}
        \draw[black!35] (\x,\ymax) -- (\x+\dx,\ymax+\dy);

    \foreach \t in {1,2}
        \draw[black!35]
        (\xmin+\t*\dx/3,\ymax+\t*\dy/3) --
        (\xmax+\t*\dx/3,\ymax+\t*\dy/3);


    \path[fill=red!35, opacity=0.80] (0,0) rectangle (3,3);

    \path[fill=red!25, opacity=0.80]
        (3,0) -- (3+\dx,0+\dy) --
        (3+\dx,3+\dy) -- (3,3) -- cycle;

    \path[fill=red!22, opacity=0.85]
        (0,3) -- (0+\dx,3+\dy) --
        (3+\dx,3+\dy) -- (3,3) -- cycle;

    \draw[black!75, dashed, thick] (0,0) rectangle (3,3);
    \draw[black!75, dashed, thick]
        (3,0) -- (3+\dx,0+\dy) --
        (3+\dx,3+\dy) -- (3,3) -- cycle;
    \draw[black!75, dashed, thick]
        (0,3) -- (0+\dx,3+\dy) --
        (3+\dx,3+\dy) -- (3,3) -- cycle;

\end{scope}

\draw[arr] (lidar.east) -- (down.west);
\draw[arr] (down.east) -- (rvfm.west);


\node[
    draw=Cenv!70,
    rounded corners=4pt,
    fill=Cenv!5,
    minimum width=1.70cm,
    minimum height=1.85cm,
    inner sep=0.5mm
] (unc) at (0.6,1.72) {};

\node[font=\scriptsize, align=center] at ($(unc.north)+(0,-0.4)$)
{Semantic\\uncertainty $H$};

\begin{scope}[shift={($(unc.center)+(0,-0.2)$)}, scale=0.16]
    \draw[->, black] (-4,-2) -- (5,-2);
    \draw[->, black] (-4,-2) -- (-4,2.9);

    \draw[blue!70, thick, smooth] plot coordinates {
        (-3.8,-1.9) (-3.0,-1.4) (-2.3,0.4) (-1.6,1.4)
        (-0.9,0.1) (-0.1,-1.3) (0.6,-1.9)};
    \fill[blue!35, opacity=0.35] (-3.8,-1.9)
        plot[smooth] coordinates {
        (-3.8,-1.9) (-3.0,-1.4) (-2.3,0.4) (-1.6,1.4)
        (-0.9,0.1) (-0.1,-1.3) (0.6,-1.9)}
        -- cycle;

    \draw[gray!80, thick, smooth] plot coordinates {
        (-2.0,-1.9) (-1.3,-1.2) (-0.7,0.8) (0,2.2)
        (0.7,0.7) (1.4,-1.1) (2.0,-1.9)};
    \fill[gray!40, opacity=0.35] (-2.0,-1.9)
        plot[smooth] coordinates {
        (-2.0,-1.9) (-1.3,-1.2) (-0.7,0.8) (0,2.2)
        (0.7,0.7) (1.4,-1.1) (2.0,-1.9)}
        -- cycle;

    \draw[red!70, thick, smooth] plot coordinates {
        (0.2,-1.9) (0.8,-1.4) (1.4,-0.1) (2.0,0.6)
        (2.8,-0.2) (3.5,-1.4) (4.0,-1.9)};
    \fill[red!35, opacity=0.35] (0.2,-1.9)
        plot[smooth] coordinates {
        (0.2,-1.9) (0.8,-1.4) (1.4,-0.1) (2.0,0.6)
        (2.8,-0.2) (3.5,-1.4) (4.0,-1.9)}
        -- cycle;

\end{scope}

\node[
    draw=Cenv!70,
    rounded corners=4pt,
    fill=Cenv!5,
    minimum width=1.70cm,
    minimum height=1.85cm,
    inner sep=0.5mm
] (akl) at (2.6,1.72) {};

\node[font=\scriptsize, align=center] at ($(akl.north)+(0,-0.4)$)
{Adaptive kernel\\length $L$};

\begin{scope}[shift={($(akl.center)+(0,-0.10)$)}, scale=0.18]
    \draw[->, thick] (-4,-2.2) -- (4,-2.2);
    \foreach \x/\r in {-3/0.45,-1.7/0.70,-0.2/1.05,1.8/1.55}
    {
        \fill[Cenv!45, opacity=0.6] (\x,0) circle (\r);
        \fill[Cenv!80!black] (\x,0) circle (0.07);
    }
    \node[font=\tiny] at (-2.6,-3.1) {Low $H$};
    \node[font=\tiny] at (2.6,-3.2) {High $H$};
\end{scope}

\node[
    draw=Cenv!70,
    rounded corners=4pt,
    fill=Cenv!5,
    minimum width=1.70cm,
    minimum height=1.85cm,
    inner sep=0.5mm
] (kinf) at (4.6,1.72) {};

\node[font=\scriptsize, align=center] at ($(kinf.north)+(0,-0.3)$)
{Kernel inference};

\begin{scope}[shift={($(kinf.center)+(0,-0.02)$)}, scale=0.15]
    \draw[black!70, thick, fill=Cenv!35] (-0.8,-0.8) rectangle (0.8,0.8);
    \draw[black!70, thick] (-0.8,0.8) -- (-0.2,1.3);
    \draw[black!70, thick] (0.8,0.8) -- (1.4,1.3);
    \draw[black!70, thick] (-0.2,1.3) -- (1.4,1.3);
    \draw[black!70, thick] (0.8,-0.8) -- (1.4,-0.3);
    \draw[black!70, thick] (1.4,-0.3) -- (1.4,1.3);

    \foreach \x/\y/\c in {
        -3/1.5/purple,
        -3/-0.4/orange,
        -1.4/-2.0/yellow,
        0.3/2.5/red,
        2.7/1.8/red,
        3.2/-0.3/pink,
        1.8/-1.9/brown,
        -0.2/-2.7/cyan}
    {
        \draw[dashed, gray!70] (0,0) -- (\x,\y);
        \fill[\c!80] (\x,\y) circle (0.18);
    }

    \node[font=\tiny, align=center] at (0,-4.5) {Kernel-weighted \\ update};
\end{scope}

\draw[arr] (unc.east) -- (akl.west);
\draw[arr] (akl.east) -- (kinf.west);


\node[
    draw=Ccritic!70,
    rounded corners=4pt,
    fill=Ccritic!5,
    minimum width=2.5cm,
    minimum height=1.6cm,
    inner sep=0.5mm
] (prev) at (1.2,-1.72) {};

\node[font=\scriptsize, align=center] at ($(prev.north)+(0,-0.26)$)
{Previous counter $\sigma_{t-1}$};

\begin{scope}[shift={($(prev.center)+(0.15,-0.20)$)}, scale=0.22]
    \draw[->, black] (-3.2,-1.2) -- (2.9,-1.2);
    \draw[->, black] (-3.2,-1.2) -- (-3.2,2.5);

    \foreach \y/\c/\w in {
        1.3/purple/2.5,
        0.75/green/2.2,
        0.2/blue/1.5,
        -0.35/orange/1.35,
        -0.9/yellow/0.95}
    {
        \fill[\c!85] (-2.8,\y) rectangle (-2.3,\y+0.26);
        \fill[gray!55] (-2.0,\y) rectangle ++(\w,0.26);
    }
\end{scope}

\node[
    draw=Ccritic!70,
    rounded corners=4pt,
    fill=Ccritic!5,
    minimum width=2.5cm,
    minimum height=1.6cm,
    inner sep=0.5mm
] (tdecay) at (4.,-1.72) {};

\node[font=\scriptsize, align=center] at ($(tdecay.north)+(0,-0.26)$)
{Temporal decay $\alpha$};

\begin{scope}[shift={($(tdecay.center)+(0.15,-0.18)$)}, scale=0.23]
    \draw[->, black] (-2.8,-1.1) -- (2.5,-1.1);
    \draw[->, black] (-2.8,-1.1) -- (-2.8,2.3);
    \node[font=\tiny] at (-3.5,1.35) {$1$};
    \node[font=\tiny] at (-3.5,-1.0) {$0$};

    \draw[red!80, very thick, smooth] plot coordinates {
        (-2.6,1.3) (-2.1,0.6) (-1.5,0.15) (-0.8,-0.12)
        (-0.1,-0.30) (0.8,-0.48) (1.6,-0.58) (2,-0.63)
    };

    \node[font=\scriptsize] at (2.1,-1.8) {$\Delta t$};
\end{scope}

\draw[arr] (prev.east) -- (tdecay.west);


\node[
    draw=Cpd!80,
    rounded corners=5pt,
    fill=Cpd!10,
    minimum width=1.2cm,
    minimum height=1.2cm,
    inner sep=1mm
] (fusion) at (6.5,0) {};

\node[font=\bfseries\scriptsize, align=center] at ($(fusion.center)+(0,0)$) {FUSION};

\node[
    draw=Cbuf!75,
    rounded corners=5pt,
    fill=Cbuf!8,
    minimum width=2.55cm,
    minimum height=3cm,
    inner sep=1mm
] (map) at (8.75,0) {};

\node[font=\bfseries\scriptsize, align=center] at ($(map.north)+(0,-0.5)$)
{Continuous\\semantic map};

\node at ($(map.center)+(0,-0.23)$)
{\includegraphics[width=2.05cm]{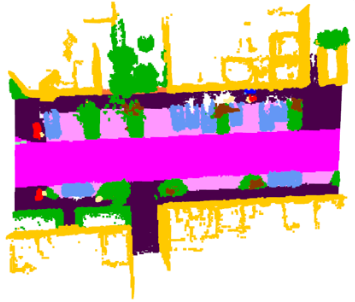}};

\coordinate (inputOut) at (-0.30,1.72);
\coordinate (spatialIn) at (0.10,1.72);
\draw[arr, rounded corners=3pt]
    (rvfm.east) -- ++(0.35,0) |- (inputOut);
\coordinate (fusionInTop) at (fusion.north);
\coordinate (fusionInBot) at (fusion.south);

\coordinate (fusionTopRoute) at ($(fusion.north)+(0,0.25)$);

\coordinate (fusionBotRoute) at ($(fusion.south)+(0,-0.25)$);

\draw[arr, rounded corners=3pt]
    (kinf.east) -- ++(0.35,0) -| (fusionTopRoute) -- (fusionInTop);

\draw[arr, rounded corners=3pt]
    (tdecay.east) -- ++(0.35,0) -| (fusionBotRoute) -- (fusionInBot);

\draw[arr] (fusion.east) -- (map.west);

\draw[arr, rounded corners=3pt]
    (map.south) -- ++(0,-1.65) -| (prev.south);

\end{tikzpicture}}
\caption{Overall architecture of the proposed spatio-temporal semantic mapping system.}
\label{fig:framework}
\end{figure*}

\section{Proposed Method}
\label{sec:method}

The proposed method builds a continuous semantic map by combining geometry, semantic labels, semantic uncertainty, and temporal history. The input is a sequence of LiDAR point clouds whose points carry 3D coordinates and a semantic class from a segmentation model. As shown in Fig.~\ref{fig:framework}, the cloud is first downsampled and mapped into a voxel grid based on an octree structure~\cite{hornung2013octomap}, and a redundant voxel filtering module discards voxels with little useful information, such as those with low point density or outside the region of interest. The semantic uncertainty of each remaining voxel is then estimated and used to set the adaptive kernel length for spatial inference. Finally, the spatial result is fused with the map from previous time steps through a temporal decay mechanism, giving a map that is both spatially smooth and temporally consistent.

\subsection{Semantic Uncertainty Estimation}

In semantic mapping, each voxel can be described by a probability distribution over semantic classes. If one class has a much higher probability than the others, the label of the voxel is reliable. If several classes have similar probabilities, the system has low confidence in the label. Such a voxel has high semantic uncertainty.
In this work, semantic uncertainty is measured using Shannon entropy:

\begin{equation}
H(v) = - \sum_{c=1}^{C} p(c|v) \log p(c|v),
\end{equation}

where $v$ is a voxel, $C$ is the number of semantic classes, $p(c|v)$ is the probability that voxel $v$ belongs to class $c$, and $H(v)$ is the semantic uncertainty of the voxel. A larger entropy means higher uncertainty. Voxels with high entropy usually appear near object boundaries, in occluded regions, in sparse-point regions, or in areas where the segmentation model confuses similar classes. These voxels need more context from their neighbors during semantic inference.

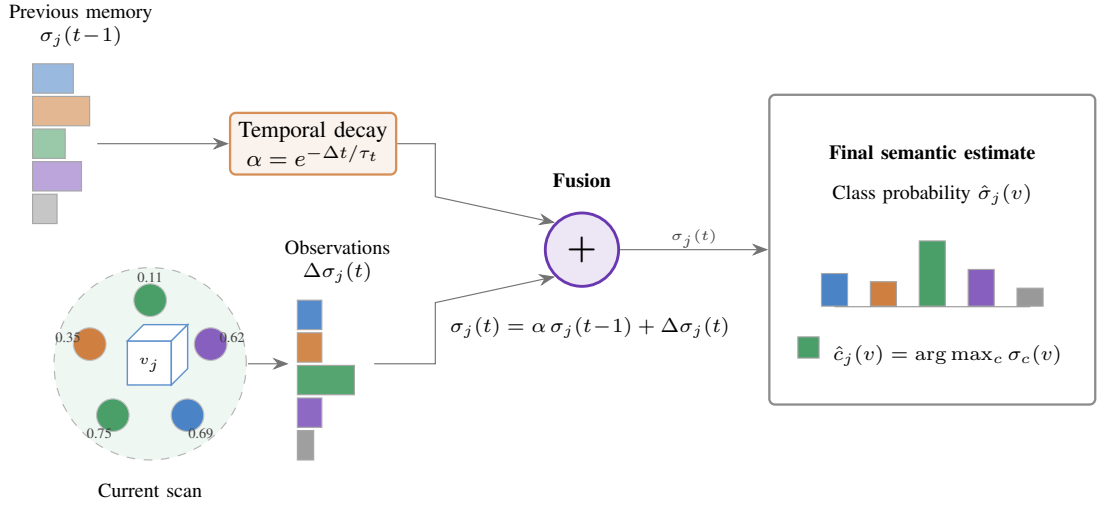
\begin{figure*}[!ht]
    \centering
    \resizebox{0.8\textwidth}{!}{%
    \begin{tikzpicture}[
      font=\footnotesize,
      >={Stealth[length=2mm]},
      cell/.style={draw=black!35, anchor=west, minimum height=3.6mm, inner sep=0pt},
      bar/.style={draw=black!35, anchor=south, minimum width=3.2mm, inner sep=0pt},
      dot/.style={circle, draw=black!25, minimum size=4mm, inner sep=0pt},
      wlbl/.style={font=\tiny, text=black!75, inner sep=0.6pt},
      capt/.style={align=center, font=\scriptsize},
      bx/.style={rounded corners=2pt, draw, thick, align=center, inner sep=3pt},
      ttl/.style={font=\scriptsize\bfseries},
    ]
    \colorlet{cA}{Cactor}\colorlet{cB}{Ccritic}\colorlet{cC}{Cenv}\colorlet{cD}{Cbuf}\colorlet{cE}{black!50}
    \foreach \i/\c/\w in {0/cA/5,1/cB/7,2/cC/4,3/cD/6,4/cE/3}{
      \node[cell, fill=\c!45, minimum width=\w mm] at (0.25, {4.5-\i*0.40}) {};
    }
    \node[capt] at (0.85,5.15) {Previous memory\\$\sigma_{j}(t\!-\!1)$};
    \node[bx, fill=Ccritic!8, draw=Ccritic!70] (decay) at (3.7,3.7) {Temporal decay\\[1pt]$\alpha=e^{-\Delta t/\tau_t}$};
    \begin{scope}[shift={(1.7,1.0)}]
      \begin{scope}[on background layer]\fill[Cenv!7] (0,0) circle (1.18);\end{scope}
      \draw[dashed, black!30] (0,0) circle (1.18);
      \begin{scope}[shift={(-0.28,-0.26)}, line width=0.5pt]
        \filldraw[fill=white, draw=Cactor!70] (0,0) rectangle (0.54,0.54);
        \filldraw[fill=white, draw=Cactor!70] (0,0.54) -- (0.2,0.74) -- (0.74,0.74) -- (0.54,0.54) -- cycle;
        \filldraw[fill=white, draw=Cactor!70] (0.54,0) -- (0.74,0.2) -- (0.74,0.74) -- (0.54,0.54) -- cycle;
        \node[font=\tiny] at (0.27,0.25) {$v_j$};
      \end{scope}
      \foreach \ang/\col/\wt in {90/cC/0.11, 162/cB/0.35, 234/cC/0.75, 306/cA/0.69, 18/cD/0.62}{
         \node[dot, fill=\col!80] at (\ang:0.78){};
         \node[wlbl] at (\ang:1.06){\wt};
      }
    \end{scope}
    \node[capt] at (1.7,-0.55) {Current scan};
    \foreach \i/\c/\w in {0/cA/3,1/cB/3,2/cC/7,3/cD/3,4/cE/2}{
      \node[cell, fill=\c!75, minimum width=\w mm] at (3.5, {1.6-\i*0.40}) {};
    }
    \node[capt] at (4.0,2.25) {Observations\\$\Delta\sigma_{j}(t)$};
    \node[circle, draw=Cbuf, fill=Cbuf!12, line width=1pt, minimum size=9mm] (fuse) at (7.0,2.4) {\large$\boldsymbol{+}$};
    \node[ttl] at (7.0,3.25) {Fusion};
    \node[capt] at (7.1,1.45) {$\sigma_{j}(t)=\alpha\,\sigma_{j}(t\!-\!1)+\Delta\sigma_{j}(t)$};
    \node[bx, draw=black!45, fit={(9.3,0.5)(13.3,4.3)}, inner sep=0pt] (fpanel) {};
    \node[ttl] at (11.3,3.6) {Final semantic estimate};
    \begin{scope}[shift={(10.7,2.30)}]
      \node[capt] at (0.6,0.78) {Class probability $\hat{\sigma}_j(v)$};
      \foreach \i/\c/\h in {0/cA/0.40,1/cB/0.30,2/cC/0.80,3/cD/0.45,4/cE/0.22}{
         \node[bar, fill=\c!80, minimum height=\h cm] at ({\i*0.60-0.60},-0.6) {};
      }
      \draw[black!40] (-0.6,-0.6) -- (1.8,-0.6);
      \node[bar, fill=cC!80, minimum height=0.26cm, minimum width=2.6mm, anchor=center] at (-0.92,-1.1) {};
      \node[anchor=west, font=\scriptsize] at (-0.74,-1.2) {$\hat{c}_j(v)=\arg\max_c \sigma_c(v)$};
    \end{scope}
    \draw[->, black!55] (1.05,3.7) -- (decay.west);
    \draw[->, black!55] (decay.east) -| (5.2,3.05) -- (fuse.north west);
    \draw[->, black!55] (2.9,1.0) -- (3.4,1.0);
    \draw[->, black!55] (4.45,1.0) -| (5.2,1.75) -- (fuse.south west);
    \draw[->, black!55] (fuse.east) -- node[wlbl,above]{$\sigma_j(t)$} (fpanel.west);
    \end{tikzpicture}}
    \caption{Illustration of semantic counter fusion. The previous counter $\sigma_j(t\!-\!1)$ is scaled by the temporal decay factor $\alpha=e^{-\Delta t/\tau_t}$ and added to the kernel-weighted observations of the current scan, $\Delta\sigma_j(t)$, giving the updated counter $\sigma_j(t)$. Normalizing $\sigma_j(t)$ produces the class probability $\hat{\sigma}_j(v)$, and its argmax gives the final label $\hat{c}_j(v)$. The number next to each point is its kernel weight, and the cell and bar sizes represent the per-class counts.}
    \label{fig:fusion}
\end{figure*}

\subsection{Adaptive Kernel Length Model}

Kernel-based inference models how semantic information from one location influences nearby voxels. The kernel length sets the spatial range over which this information spreads. Using a single fixed kernel length for the whole map is not a good choice. If it is too small, the system cannot smooth noisy regions. If it is too large, semantic information is over-smoothed, object boundaries become blurred, and the computational cost grows.
To handle this, the proposed method adapts the kernel length to the semantic uncertainty of each voxel. Let $\bar{H}(v) \in [0,1]$ be the entropy $H(v)$ normalized by its maximum value. The kernel length is set by linear interpolation between a minimum and a maximum value:

\begin{equation}
L(v) = L_{\min} + (L_{\max} - L_{\min})\,\bar{H}(v),
\end{equation}
where $L_{\min}$ and $L_{\max}$ are the smallest and largest allowed kernel lengths. In uncertain regions, such as object boundaries or moving objects, the kernel length grows so that more neighboring information is used to correct the label. In stable regions, such as roads and walls, the kernel length shrinks to save computation and keep boundaries sharp. This bounded form keeps the kernel length within a safe range and removes the need to tune a single fixed value for the whole map. 

The spatial weight between a voxel $v_j$ and an observed point $p_i$ is computed with a sparse kernel that has compact support~\cite{melkumyan2009sparse, deng2024seecsom}. Let $d = \|x_j - x_i\|$ be the Euclidean distance between the voxel center and the point, and let $l = L(v_j)$ be the adaptive kernel length. The kernel is

\begin{equation}
k(v_j, p_i) =
\begin{cases}
\sigma_0\, g\!\left(\dfrac{d}{l}\right), & d < l, \\[6pt]
0, & d \ge l,
\end{cases}
\label{eq:sparse_kernel}
\end{equation}
where $\sigma_0$ is a positive scaling constant and the shape function $g(\cdot)$ is

\begin{equation}
g(r) = \frac{2 + \cos(2\pi r)}{3}\,(1 - r) + \frac{\sin(2\pi r)}{2\pi}.
\end{equation}

The weight is largest for points close to the voxel and decreases smoothly to zero at the kernel boundary ($d = l$). Unlike a Gaussian kernel, this kernel has compact support, so points beyond the kernel length have no effect. This keeps the inference local and efficient.

\subsection{Spatial Semantic Inference}

After the kernel length of each voxel is set, the semantic state of the voxel is updated from the nearby semantic points using a Bayesian kernel update. The semantic state is stored as a per-class counter. For a voxel $v_j$, the contribution of each observed point $p_i$ is weighted by the kernel in \eqref{eq:sparse_kernel}, and the counter from the current scan is accumulated as:

\begin{equation}
\Delta\sigma_t(v_j) = \sum_{i} k(v_j, p_i)\, \mathbf{1}_{c_i},
\end{equation}
where $\mathbf{1}_{c_i}$ is the one-hot vector of the observed label $c_i$. Points close to the voxel contribute more, points near the kernel boundary contribute less, and points beyond the kernel length do not contribute at all. The counter $\Delta\sigma_t(v_j)$ is the new observation that is passed to the temporal fusion step.

\subsection{Temporal Semantic Fusion}

After spatial inference, temporal fusion keeps the labels consistent across consecutive observations. During robot motion, the same region may be observed several times. Because of viewpoint changes, sensor noise, and segmentation errors, the label of the same voxel can change over time. If only the current observation is used, the map becomes unstable. If all past observations are added without considering how recent they are, outdated or wrong labels keep influencing the final map.
To solve this, the proposed method keeps a semantic counter for each voxel over time. The counter stores the accumulated evidence for each class from past observations. When a new scan arrives, the previous counter is first decayed by a temporal factor and then combined with the counter from the current scan:

\begin{equation}
\sigma_t(v) = \alpha\,\sigma_{t-1}(v) + \Delta\sigma_t(v),
\qquad \alpha = \exp\!\left(-\frac{\Delta t}{\tau_t}\right),
\label{eq:temporal_fusion}
\end{equation}
where $\sigma_t(v)$ is the fused counter at time $t$, $\sigma_{t-1}(v)$ is the counter from the previous step, $\Delta\sigma_t(v)$ is the new counter from the current scan, $\Delta t$ is the time gap since the voxel was last updated, and $\tau_t$ controls how long past information is kept.
 
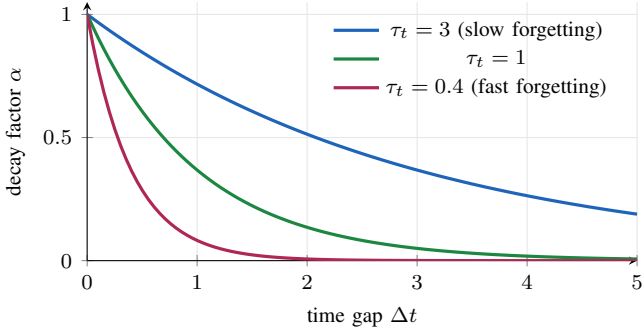
\begin{figure}[!h]
    \centering
    \begin{tikzpicture}
    \begin{axis}[
      width=\linewidth, height=5cm,
      xlabel={time gap $\Delta t$}, ylabel={decay factor $\alpha$},
      xmin=0, xmax=5, ymin=0, ymax=1.05,
      legend pos=north east, legend style={font=\footnotesize, draw=none, fill=none},
      tick label style={font=\footnotesize}, label style={font=\footnotesize},
      axis lines=left, grid=both, grid style={black!10},
    ]
    \addplot[Cactor, very thick, domain=0:5, samples=100] {exp(-x/3)};   \addlegendentry{$\tau_t=3$ (slow forgetting)}
    \addplot[Cenv, very thick, domain=0:5, samples=100] {exp(-x/1)};     \addlegendentry{$\tau_t=1$}
    \addplot[Cpd, very thick, domain=0:5, samples=100] {exp(-x/0.4)};    \addlegendentry{$\tau_t=0.4$ (fast forgetting)}
    \end{axis}
    \end{tikzpicture}
    \caption{Effect of $\tau_t$ on the temporal decay factor $\alpha = \exp(-\Delta t/\tau_t)$. A larger $\tau_t$ keeps old information longer, while a smaller $\tau_t$ forgets it faster.}
    \label{fig:alpha}
\end{figure}

A larger $\tau_t$ makes $\alpha$ close to $1$, so more historical information is kept and the map is more stable in static regions. A smaller $\tau_t$ makes $\alpha$ close to $0$, so the map forgets old observations quickly and adapts faster to changes. Using a time-dependent factor instead of a constant decay lets the decay depend on the real time gap between observations, so a scan that arrives after a longer interval loses more weight. This effect is shown in Fig.~\ref{fig:alpha}.
The class probability of each voxel is obtained by normalizing the fused counter:

\begin{equation}
\lambda^{(k)}(v) = \frac{\sigma_t^{(k)}(v)}{\sum_{h} \sigma_t^{(h)}(v)},
\end{equation}
and the final semantic label is the class with the highest counter:

\begin{equation}
\hat{c}_t(v) = \argmax_c \sigma_t^{c}(v).
\end{equation}

This mechanism keeps the map from changing abruptly because of occasional wrong predictions, while still letting it adapt when the environment truly changes. By combining adaptive spatial inference with temporal fusion, the proposed method produces a semantic map that is more accurate, less noisy, and more stable than spatial-only mapping methods.

\section{Experimental Results}
\label{sec:exp}

\subsection{Experimental Setup}

\begin{figure}[!h]
    \centering
    \includegraphics[width=1\linewidth]{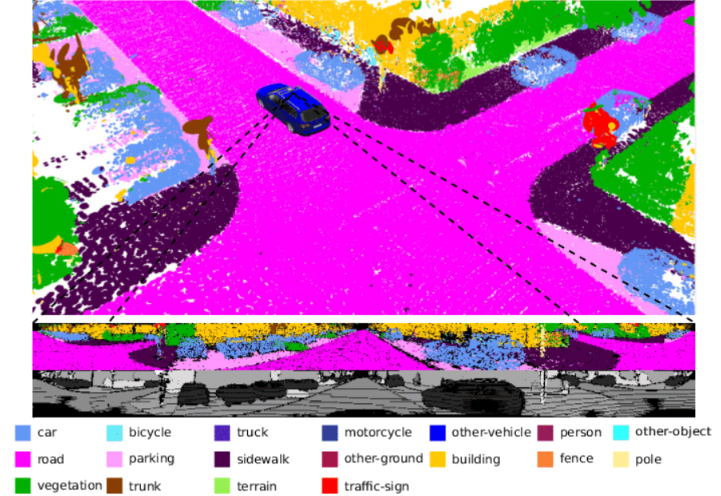}
    \caption{The SemanticKITTI dataset~\cite{behley2019semantickitti}.}
    \label{fig:dataset1}
\end{figure}

The proposed method is evaluated on the SemanticKITTI dataset~\cite{behley2019semantickitti}, which is a widely used outdoor dataset for autonomous driving perception. It contains 3D LiDAR point clouds collected from real traffic scenarios, with pointwise semantic annotations. The dataset includes many semantic classes, such as road, sidewalk, car, pedestrian, pole, and traffic sign, as shown in Fig.~\ref{fig:dataset1}. 

To evaluate the quality of semantic maps, four metrics are used, including per-class Intersection over Union (IoU), mean Intersection over Union (mIoU), weighted-average IoU (wIOU), and Overall Accuracy (OA).
For each semantic class $c$, IOU is defined as:

\begin{equation}
IoU_c = \frac{TP_c}{TP_c + FP_c + FN_c},
\end{equation}
where $TP_c$, $FP_c$, and $FN_c$ are the numbers of true positive, false positive, and false negative voxels of class $c$. The mean IoU is the average IoU over all classes:

\begin{equation}
mIoU = \frac{1}{C}\sum_{c=1}^{C} IoU_c,
\end{equation}
where $C$ is the number of evaluated classes. Because SemanticKITTI has a highly imbalanced class distribution, we also report the weighted-average IoU:

\begin{equation}
wIoU = \frac{\sum_{c=1}^{C} IoU_c \cdot n_c}{\sum_{c=1}^{C} n_c},
\end{equation}
where $n_c$ is the number of evaluated voxels of class $c$. Overall Accuracy is defined as:

\begin{equation}
OA = \frac{\sum_{c=1}^{C} TP_c}{\sum_{c=1}^{C} (TP_c + FP_c + FN_c)}.
\end{equation}

Among these metrics, mIoU is the main criterion because it reflects the average mapping performance over all classes. The wIOU and OA are reported as complementary metrics that show the overall map quality under class imbalance.

\subsection{Effect of Adaptive Kernel Length}

\begin{table}[!h]
\centering
\caption{Evaluate the effect of kernel length.}
\label{tab:kernel_length}
\begin{tabularx}{\linewidth}{
   >{\centering\arraybackslash}X||
   >{\centering\arraybackslash}m{0.15\textwidth}
   >{\centering\arraybackslash}X}
\toprule
Kernel Length & Runtime (10 scans) (s) & Accuracy (\%) \\
\midrule
1 & 40.86 & \second{73.02} \\
2 & \second{40.48} & 72.80 \\
4 & 47.31 & 72.66 \\
6 & 49.90 & 72.61 \\
Adaptive & \best{39.61} & \best{78.16} \\
\bottomrule
\end{tabularx}
\end{table}

\begin{table}[!t]
\centering
\caption{Per-class IoU (\%), OA (\%), and wIoU (\%) on the SemanticKITTI dataset. For each column within a sequence, the \best{best} result is shown in purple and the \second{second-best} in bold.}
\label{tab:multiseq}
\scriptsize
\setlength{\tabcolsep}{2pt}
\begin{tabularx}{\columnwidth}{
   >{\centering\arraybackslash}m{0.02\textwidth}
   >{\raggedright\arraybackslash}m{0.063\textwidth}
   *{10}{>{\centering\arraybackslash}X}}

seq. & \multicolumn{1}{l}{\bf Method}
& \cellcolor{sbicycleColor}\rotatebox{90}{\color{white}Bicycle}
& \cellcolor{spersonColor}\rotatebox{90}{\color{white}Person}
& \cellcolor{sparkingColor}\rotatebox{90}{\color{white}Parking}
& \cellcolor{ssidewalkColor}\rotatebox{90}{\color{white}Sidewalk}
& \cellcolor{sothergroundColor}\rotatebox{90}{\color{white}Other Ground}
& \cellcolor{sbuildingColor}\rotatebox{90}{\color{white}Building}
& \cellcolor{strunkColor}\rotatebox{90}{\color{white}Trunk}
& \cellcolor{spoleColor}\rotatebox{90}{\color{white}Pole}
& \multicolumn{1}{c}{\bf OA} & \multicolumn{1}{c}{\bf wIoU} \\
\hline
\multirow{3}{*}{03}
 & CSM\cite{gan2020bayesian}   & 46.04 & $-$ & 82.59 & 59.93 & 68.88 & $-$ & 66.50 & \second{43.57} & \second{72.48} & 55.20 \\
 & S-BKI\cite{gan2020bayesian} & \second{51.37} & $-$ & \second{83.15} & \second{61.40} & \second{69.03} & $-$ & \second{66.85} & 42.95 & 71.97 & \second{56.38} \\
 & Ours & \best{68.61} & $-$ & \best{88.12} & \best{65.09} & \best{79.89} & $-$ & \best{77.03} & \best{53.50} & \best{81.73} & \best{69.99} \\
\midrule
\multirow{3}{*}{04}
 & CSM\cite{gan2020bayesian}   & 18.06 & 48.54 & 87.23 & 52.43 & \second{59.11} & 76.76 & 60.13 & \best{62.05} & 63.50 & 46.12 \\
 & S-BKI\cite{gan2020bayesian} & \second{20.33} & \second{49.25} & \second{88.20} & \second{53.42} & 54.49 & \best{79.71} & \second{63.50} & 53.04 & \second{64.18} & \second{49.24} \\
 & \textbf{Ours} & \best{23.01} & \best{53.40} & \best{91.03} & \best{60.48} & \best{60.50} & \second{78.17} & \best{65.29} & \second{60.90} & \best{70.10} & \best{54.28} \\
\midrule
\multirow{3}{*}{05}
 & CSM\cite{gan2020bayesian}   & 42.40 & 26.23 & 85.10 & \second{52.20} & \second{66.24} & 66.90 & 50.10 & 55.02 & 68.52 & 53.51 \\
 & S-BKI\cite{gan2020bayesian} & \second{45.08} & \second{26.66} & \second{85.43} & \best{56.01} & 65.07 & \second{67.67} & \second{55.50} & \best{76.00} & \second{71.09} & \second{55.81} \\
 & \textbf{Ours} & \best{51.91} & \best{28.29} & \best{89.71} & 43.12 & \best{72.77} & \best{69.72} & \best{59.34} & \second{75.35} & \best{75.96} & \best{61.51} \\
\midrule
\multirow{3}{*}{06}
 & CSM\cite{gan2020bayesian}   & 43.80 & 24.20 & 78.52 & 69.47 & 68.19 & 0.00 & \second{56.63} & \second{75.40} & 64.10 & 48.09 \\
 & S-BKI\cite{gan2020bayesian} & \second{47.11} & \second{25.09} & \second{78.70} & \second{69.50} & \second{68.41} & 0.00 & 56.31 & 75.33 & \second{65.07} & \second{48.31} \\
 & \textbf{Ours} & \best{51.03} & \best{36.17} & \best{80.17} & \best{71.94} & \best{71.13} & 0.00 & \best{61.74} & \best{79.51} & \best{69.30} & \best{53.20} \\
\bottomrule
\end{tabularx}
\end{table}

Table~\ref{tab:kernel_length} reports the effect of different kernel length settings on runtime and mapping accuracy. Fixed kernel lengths of 1, 2, 4, and 6 are compared with the proposed adaptive kernel length.
The results show that fixed kernel lengths do not give a good balance between accuracy and computational cost. As the kernel length grows from 1 to 6, the runtime increases from 40.86~s to 49.90~s, while the accuracy slightly drops from 73.02\% to 72.61\%. This means that a larger fixed kernel can include irrelevant neighboring voxels and spread semantic noise, especially near object boundaries. In contrast, the adaptive kernel length gives the best performance, with the highest accuracy of 78.16\% and the lowest runtime of 39.61~s. 

Compared with the best fixed-kernel setting, the adaptive strategy improves accuracy by 5.14 percentage points and is also faster. This confirms that adapting the kernel length to semantic uncertainty improves accuracy while avoiding unnecessary computation.

\subsection{Quantitative results}
To evaluate the robustness of the proposed method, experiments are run on four SemanticKITTI sequences: 03, 04, 05, and 06. The compared methods are S-CSM\cite{gan2020bayesian}, S-BKI\cite{gan2020bayesian}, and the proposed spatio-temporal semantic mapping method. S-CSM accumulates semantic observations with a count-and-smooth strategy, and S-BKI applies Bayesian kernel inference with a fixed kernel. The proposed spatio-temporal mapping method is compared with CSM and BKI on several representative semantic classes, together with Overall Accuracy (OA) and weighted-average IoU (wIOU). 

As shown in Table~\ref{tab:multiseq}, the proposed method consistently achieves the best OA and wIOU on all evaluated sequences. Compared with CSM, the proposed method improves OA by 9.25, 6.60, 7.44, and 5.20 percentage points on sequences 03, 04, 05, and 06. The corresponding gains in wIOU are 14.79, 8.16, 8.00, and 5.11 percentage points. These results indicate that the proposed spatio-temporal fusion strategy is not limited to one sequence, but stays effective across different outdoor scenes. The improvement is mainly due to the combination of adaptive spatial inference and temporal decay. Spatial inference improves local semantic consistency, while temporal fusion stabilizes labels across consecutive scans and reduces the effect of noisy single-frame predictions.

\begin{table}[!h]
\centering
\caption{Ablation Study on KITTI Sequence 03.}
\label{tab:ablation}
\begin{tabularx}{\linewidth}{
   >{\raggedright\arraybackslash}m{0.13\textwidth}||
   >{\centering\arraybackslash}X
   >{\centering\arraybackslash}m{0.1\textwidth}
   >{\centering\arraybackslash}X}
\toprule
Method & wIOU (\%) & OA (\%) & mIoU (\%) \\
\midrule
w/o temporal & 64.30 & 78.23 & 41.74 \\
Proposed Method & \best{69.99} & \best{81.73} & \best{54.92} \\
\bottomrule
\end{tabularx}
\end{table}

Table~\ref{tab:ablation} shows the ablation study with spatial semantic mapping, which uses semantic uncertainty for adaptive spatial inference but does not use temporal fusion. The proposed method improves wIOU from 64.30\% to 69.79\%, accuracy from 78.23\% to 81.61\%, and mIoU from 41.74\% to 54.92\%. The 13.18 percentage-point gain in mIoU shows the value of temporal semantic fusion. The improvement comes from the complementary roles of spatial and temporal inference. Adaptive spatial inference improves local semantic consistency by using semantic uncertainty to control the kernel range. However, spatial inference alone cannot handle inconsistent predictions across consecutive scans. By adding temporal decay, the proposed method reduces the effect of outdated semantic information while keeping useful past observations. As a result, the map becomes more robust to occasional segmentation errors and semantic drift.

\subsection{Qualitative Evaluation}
\begin{figure}[!t]
    \centering
    \includegraphics[width=\linewidth]{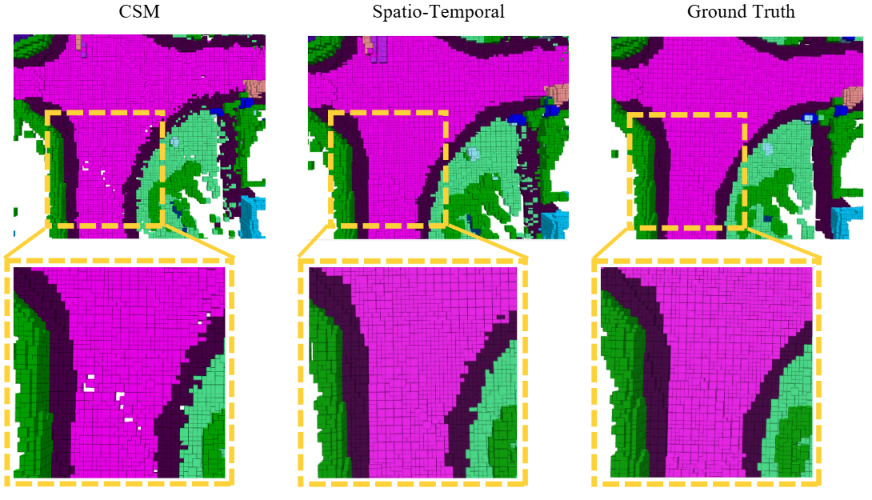}
    \caption{Qualitative comparison of semantic maps on SemanticKITTI sequence 05.}
    \label{fig:qualitative_seq05}
\end{figure}

\begin{figure}[!ht]
    \centering
    \includegraphics[width=\linewidth]{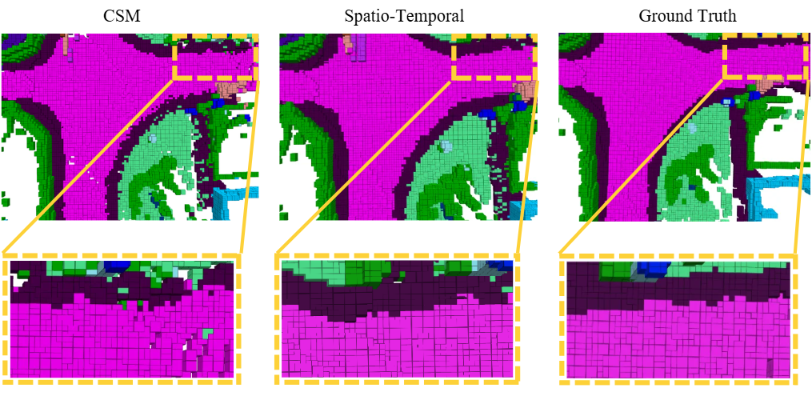}
    \caption{Qualitative comparison of semantic maps on SemanticKITTI sequence 06.}
    \label{fig:qualitative_seq06}
\end{figure}
In addition to the numbers, qualitative results are used to study the visual quality of the maps. Fig.~\ref{fig:qualitative_seq05} shows a visual comparison of the SemanticKITTI sequence 05.
The baseline result still contains many isolated misclassified voxels and discontinuous semantic regions. These errors are mainly caused by noisy single-frame predictions, sparse LiDAR observations, and the lack of temporal consistency between consecutive scans. Spatial smoothing can reduce some of the local noise, but inconsistent labels can still remain near object boundaries or in regions with little geometric information. 

In contrast, the proposed spatio-temporal method produces a cleaner and more consistent map. Large regions such as roads, vegetation, and buildings are better preserved, while isolated noisy labels are reduced. The temporal fusion mechanism stabilizes the labels when the same region is observed several times, and the adaptive spatial inference improves local consistency near uncertain regions. As a result, the map becomes visually closer to the ground truth and more suitable for downstream navigation tasks.

Similar improvements can also be seen on sequence 06, as shown in Fig.~\ref{fig:qualitative_seq06}. The proposed method reduces fragmented regions and improves the continuity of the semantic structures compared with the baseline method.

\section{Conclusion}
\label{sec:conclusion}

This paper presented a continuous semantic mapping method that combines spatial and temporal information for autonomous robots. The method estimates the semantic uncertainty of each voxel with Shannon entropy and uses this uncertainty to adapt the kernel length during spatial inference. It also applies a time-aware decay mechanism that gives more weight to recent observations while keeping useful past information. Experimental results across several SemanticKITTI sequences confirm that using spatial and temporal reasoning together makes the semantic map more accurate, less noisy, and more stable over time. However, the method still has limitations. It depends on the quality of the input segmentation, and it does not yet separate moving objects from static structures, so errors can still accumulate in dynamic regions. Future work will integrate stronger LiDAR segmentation networks, add motion segmentation to handle dynamic objects, and evaluate the semantic map in real and simulated navigation tasks. Extending the method to multi-robot mapping is another promising direction.


\bibliographystyle{IEEEtran}
\bibliography{ref}  

\end{document}